\documentclass[runningheads]{llncs}
\usepackage[T1]{fontenc}
\usepackage{graphicx}
\begin{document}
\title{From Manifestations to Cognitive Architectures:\\ a Scalable Framework}
%
%
\author{Alfredo Ibias\inst{1}\orcidID{0000-0002-3122-4272} \and
Guillem Ramirez-Miranda\inst{1}\orcidID{0000-0003-2741-3705} \and
Enric Guinovart\inst{1} \and
Eduard Alarcon\inst{2}}
\authorrunning{A. Ibias et al.}
%
\institute{Avatar Cognition, Barcelona, Spain\\
\email{\{alfredo, guillem, enric\}@avatarcognition.com} \and
Universitat Politècnica de Catalunya - BarcelonaTech, Barcelona, Spain\\
\email{eduard.alarcon@upc.edu}}
\maketitle              
\begin{abstract}
The Artificial Intelligence field is flooded with optimisation methods. In this paper, we change the focus to developing modelling methods with the aim of getting us closer to Artificial General Intelligence. To do so, we propose a novel way to interpret reality as an information source, that is later translated into a computational framework able to capture and represent such information. This framework is able to build elements of classical cognitive architectures, like Long Term Memory and Working Memory, starting from a simple primitive that only processes Spatial Distributed Representations. Moreover, it achieves such level of verticality in a seamless scalable hierarchical way.

\end{abstract}
\section{Introduction}
Artificial Intelligence is a research field that has attracted a lot of attention lately due to its impressive results mimicking human intelligence and problem solving. These results have been accomplished by a plethora of methods and algorithms whose main goal is to approximate a target function, and they do so by reducing a loss function with respect to that ideal target function. Thus, they intrinsically become optimisation methods trying to reduce a loss function.

However, most theories about how the brain works reject the idea that the brain is an optimisation machine. Moreover, many of them propose that the brain models the world~\cite{hb04,yhp20}. This assumption can be explained with a very simple fact: an optimisation algorithm can only store one answer for each given input, the optimal answer. It does not have the information to be able to consider another options, even less to reason. Thus, it will only provide the answer it has ``stored'' as the best possible answer to the given input. In contrast, a modelling algorithm can store the information of what are the possible answers for a given input, and therefore has the potential to consider the different possible options, and even to reason about which one is the best.

This mismatch between Artificial Intelligence practice and brain theories posses a fundamental question for those that aim to achieve Artificial General Intelligence: how to transform those optimisation algorithms into modelling algorithms. There has been some attempts to do so by building complex architectures~\cite{hs18,zxychu23}. However, we argue that such a change cannot be performed, as the problem is more fundamental: it relies in the fact that these algorithms have been developed to look for a target function that solves the given problem. And the one fact about functions its that they assign only one answer for any given input. Thus, if we aim to develop an artificial intelligence that works similar to the brain, we need to look elsewhere for a starting point.

A fact about modelling the world is that the world cannot be observed all at once, but instead it has to be experienced. This poses a huge limitation to how we can model things, as we do not have the whole information of any object, but instead slices of it called \emph{manifestations}. For example, when we observe a cup, we do not observe the whole cup, neither the concept of cup, but instead we observe a perspective of an specific cup. This perspective is a manifestation of the cup in the specific conditions it is right now, and thus it can change with different conditions. For example, if you change the point of view and rotate around the cup, you will see different manifestations of it, and if you dim the lights you will see also a different manifestation of the cup (this time a darker one). However, in all these cases, a cognitive agent (i.e. a human) can still recognise the same cup, and thus can work out that all those manifestations of the same cup can be abstracted into the concept of a cup.
This \emph{abstraction} of a cup can be build because all the manifestations we observe from it come from the same object, that is, they are different manifestations of the same reality. When we look at it from an information perspective (analysing the world as a source of information), we say that all the manifestations come from the same \emph{latent information}, that is, from the same piece of information that is latent in all the manifestations. And this latent information is expressed differently based on surrounding conditions, hence the different manifestations. This latent information is what we model in the abstraction we do of the object.

If we change the focus from a lot of different manifestations that collectively represent a latent information, to a latent information that is expressed differently based on surrounding conditions, we can observe that this latent information has a fundamental property: it is persistent and self-projecting, in the sense that the latent information is always there and it is always providing a manifestation of itself, thus it cannot disappear. This property is what we call the \emph{Self-Projecting Persistence Principle}.

The Self-Projecting Persistence Principle (SPPP) states that any latent information present in the world persists in time and is always manifesting itself (based on surrounding conditions). The persistence in time is based on the fact that objects do not disappear or change, but they stay in the same state unless actuated by external forces. Thus, their latent information do not change unless external forces modify it, what in turn changes the nature of the latent information. For example, following the cup example, an empty cup tends to stay empty and its latent information is that of an empty cup, until someone comes and fills it. Then, the latent information changes to be the information of a filled cup, and thus it has transformed into a different kind of latent information. It is true that both are latent information of a cup, but they are slightly different, and their similarities correspond to the hierarchical nature of latent information.
The self-projecting part of the principle is a bit more nuanced, as it can be easily confused with a subjective point of view. The idea is that any latent information is always expressing itself to the world. We, cognitive agents, only capture one of those manifestations at any given time, but the latent information is always expressing them, in all directions. Following the cup example, a cup is always reflecting light, and therefore is always expressing itself to any sensor able to perceive light. But vision is not the only way: a cup is always expressing also its physical place in the world, in the sense that you can touch it and feel it (even in the darkness) thanks to your touch sensors.
We are aware these examples show a convoluted way to explain the physical forces of the world. However, as our aim is to extract information from the world, we need to accept this increased complexity in order to develop an information-based approach.
In essence, the SPPP principle tells us that any information that exists in the world persists in time and continually reveals itself to the world. This is a fundamental property that we will use as base for our proposal of representation.

To perceive this latent information and thus build abstractions, any agent needs an embodiment. This is based on the fact that there is no omniscient agent in the world, that can directly observe the latent information. Instead, all agents are limited by the locality of their bodies and their sensors. Thus, defining an agent's embodiment defines which kind of manifestations it can capture, and thus which kind of abstractions it can build. For example, an agent without vision cannot build the concept of colours, at least not in the same sense than an agent with it. Agents with different embodiments perceive different manifestations of the world, and thus build different abstractions of it. This in turn provides the agent's cognition mechanism with different pieces to work with, influencing the kind of reasoning an agent can perform, and therefore limiting the kind of behaviour it can develop.

Regarding the hierarchical nature of latent information, we know that any latent information contains different levels of information: for an individual cup, the main latent information is the information about that specific cup. However, it also has information about the concept of cup, i.e. about what makes a cup a cup. But moreover, it also has information about any collection that includes cups, like tableware, as we build this kind of concepts from the experiencing of their individual components. It contains information about any concept that builds the cup too, like the material it is made of, or its different parts.

All these levels of information show us that latent information, and thus abstractions, are organised in hierarchies: from the concrete, simpler concepts to the generic, more complex ones. This hierarchical nature is fundamental to how knowledge representation should work: we need to capture such hierarchies in order to fully grasp the concepts and their relationships.

Extending the hierarchical nature of abstractions, we propose, as a working definition, to group it into four main groups, namely the \emph{perception abstractions}, the \emph{sensorimotor abstractions}, the \emph{episodic abstractions} and the \emph{reasoning abstractions}. Of these four groups, only the perception abstractions can be built by any cognitive agent (with the same set of sensors), and thus it is the only information present in the latent information. The other three groups build over the perception abstractions and depend on the capabilities of the cognitive agent.

The perception abstractions are those that can be build based only on the perceived manifestations of the world around. For example, in the cup example, this kind of abstractions would be the abstraction of a concrete cup, the abstraction of the concept of cup, or the abstraction of the material the cup is made of. In the end, these are abstractions that anyone that can perceive a cup can make. These would be the kind of abstractions any animal with a brain does.

The sensorimotor abstractions are a bit more complicated, because they require the cognitive agent to interact with the world. In our preceding example, these are abstractions like how to push the cup, how to move away from the cup if it is hot, etc... In the end, they are abstractions that associate the perception of the world to actions you can make over it, and thus requires a body with a motor part. It is important to remark that this kind of abstractions do not consider time, but only instantaneous responses to instantaneous inputs, and thus this kind of abstractions are the ones reflexes are based on.

The episodic abstractions require processing not only instantaneous manifestations of the latent information, but also their evolution through time. To this group belong abstractions like how a cup falls to the floor, how to drink from a cup, etc... The key fact here is that they consider time, and therefore can produce different behaviours for the same input based on time context. For example, they allow to tell apart a still cup and a moving cup, and allow the agent to block the moving cup to avoid it falling from a table. This is the kind of abstractions mostly present in animals.

Finally, the reasoning abstractions require processing also how the brain thinks. This kind of abstractions are more convoluted and mostly present in mammals, but they are the most powerful. They abstract planning and reasoning, and thus can place goals and modify behaviour based on them. In our example, these would be abstractions like filling the cup to be able to drink from it, or the concept of how cups are used, what is the right way to pick them, etc... These abstractions are influenced not only by the individual embodiment, but also by how the individual has learned to reason.

\section{A Computational Framework of Cognition}
Based on the description we proposed in the previous Section, we developed a computational framework whose goal is to mimic the latent information description presented in the previous section. It requires to start with the most fundamental element: a commonality. An the first question is how to represent (and store) a commonality. Here appears a fundamental requisite for our representations to be able to represent any commonality: they need to be universal, i.e. input-agnostic.
Recent research has shown that a valid universal data structure is a Sparse Distributed Representation~\cite{ah16,cah17} (SDR), which allows an universal representation of the inputs independently of their type. This has been proven to be the actual way in which the brain processes its inputs~\cite{cah17,foldiak03}. Thus, our representations will be based on SDRs. However, SDRs are not the only option possible, and alternative data structures can be proposed, specially for specific goals.
Now, the use of SDRs imply that we need our embodiment to transform the signals it collects from the world into SDRs, thus needing some kind of encoder, like one that transforms images to SDRs. To be able to observe the internal representations of our computational framework, we also developed decoders that transform SDRs into samples of the original signals. These decoders are the ones used to show the representations present in the Figures of this Section in a human readable format.

\begin{figure}[tp]
    \centering
    \includegraphics[width=0.2\columnwidth]{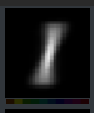}
    \includegraphics[width=0.65\columnwidth]{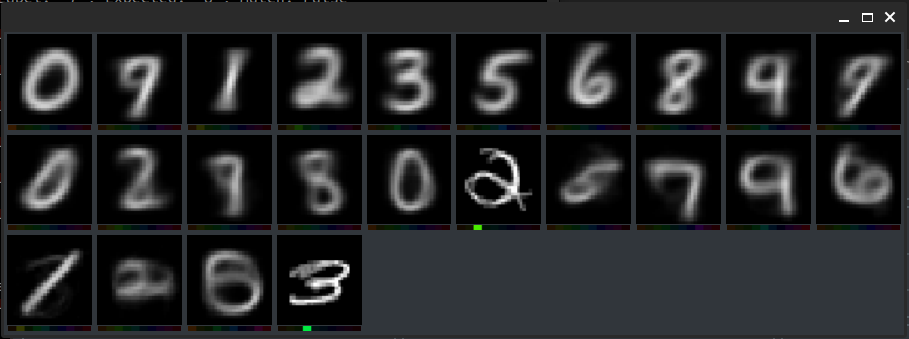}
    \caption{(left) An example of Footprint: the combination of the 1's of the MNIST dataset.\\(right) An example of Cell: the Footprints of the 60,000 samples of the MNIST dataset.}
    \label{fig:FP}
\end{figure}

Getting back to commonalities, we need to remember that in the real world they are latent information, but we cannot observe them directly. Instead, we observe manifestations of them and our brains take those manifestations and combine them in a way that a commonality is developed.
Thus, the first element we need to introduce is a method to build abstractions from concrete manifestations. This method is what we have called Footprinting, and consists in combining those concrete manifestations into the same representation. In our work we call the representations \emph{Footprints} (hence the name Footprinting for the method), and they store the abstraction of a set of concrete manifestations. An example of Footprint can be observed on the left of Figure~\ref{fig:FP}.


As explained in the previous Section, we will want to extend the Self-Projecting Persistence Principle (SPPP) to our representations (Footprints)
, and thus, they need to persist and self-project. The persistence part is easy, as we are storing them in memory. The self-projecting part is a bit more nuanced, as it first needs to define what does that mean for a representation. In our interpretation of the principle, we consider that the self-projecting part of a representation means that, when provided with a new input (i.e. a new manifestation to combine), it will output itself, that is, the combination it has built. This output is what we call the Footprint's \emph{Projection}. And it will be processed by the embodiment's decoder to produce an output to the world. Here it is important to remark that a commonality, a Footprint and a Projection share the same content. However, the commonality is such content as existing in the world, the Footprint is such content as stored in an implementation of our framework, and the Projection is such content as the output of our framework. 

Getting back to combining concrete manifestations, there are multiple possible methods to do it, and at this point we do not have any reason to say one is better than other. Right now, we decided to perform a simple averaging over all the manifestations combined into the same Footprint, but other methods could also work. Additionally, to be able to combine multiple manifestations of the same commonality into the same Footprint, we need a way to determine which manifestations come from it, and which ones do not. To that end, a similarity function should be defined that would tell how probable is that two manifestations come from the same commonality. And we also need a threshold to determine from which level of similarity is acceptable to join two manifestations.

With these two tools, we can go one step further and develop Footprints of multiple commonalities. To be consistent, we should group together those Footprints and use the same threshold for all of them. Thus, we define the concept of \emph{Cell} as the grouping of multiple Footprints that share the same threshold. An example of Cell is displayed at the right of Figure~\ref{fig:FP}.

The Cell is the basic logical level where the Footprinting is executed in its full: when a new manifestation comes, all the Footprints compute their similarity and check if they are over the Cell's threshold. If one or more similarities are over, the manifestation is assigned to the Footprint with higher similarity. If no similarity is over the threshold, then a new Footprint is needed.

When we translate the SPPP principle to the Cell, we observe that it is not very different from how it affects Footprints. It makes the Cell to provide an output that, in this case, is the Projection of the last updated Footprint. This transforms the Cell into a pattern matching function, that receives inputs, computes their similarity with respect to the patterns it has already found, and provides the most similar one. Here it is crucial to remark that, providing the pattern, you are filling the gaps. For example, if you receive half the image of a cup, the pattern matching will match with the representation of the cup, and provide the full image of a cup as a Projection.

\begin{figure}[tp]
    \centering
    \includegraphics[width=0.45\columnwidth]{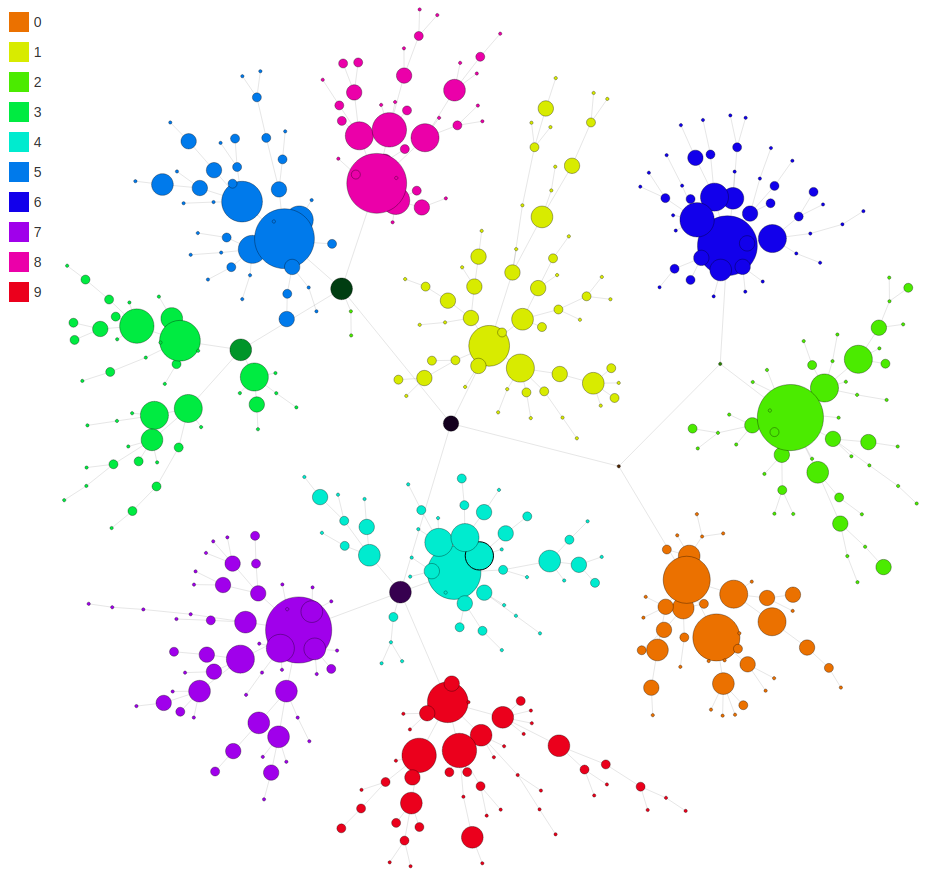}
    \includegraphics[width=0.4\columnwidth]{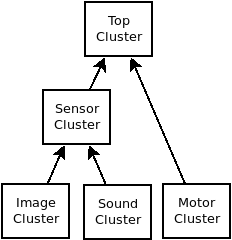}
    \caption{(left) An example of Cluster: the Cells of the hierarchy present in the 60,000 samples of the MNIST dataset. The seed Cell is the central black node.\\(right) An example of Metacluster: a Metacluster for an agent that has two sensor inputs (image and sound) and a motor output.}
    \label{fig:Cluster}
\end{figure}

The next step is to add hierarchy to our framework. As abstractions come in hierarchies (as explained before), our internal representation of them should also organise as a hierarchy. Thus, we need to build hierarchies of Cells, where the child Cells represent more specific commonalities than their parent ones, and are associated with a specific Footprint of the parent Cell. In mathematical terms, we would say that a Footprint represents a domain, and the Footprints in a child Cell are subdomains of the parent Footprint domain. With this setup, a new Cell should be created only when a Footprint represents a large enough domain.
A hierarchy of Cells is what we call a \emph{Cluster}. Each Cluster contains in fact a tree, with a seed Cell where the most generic commonalities are stored, and many branches and leaf Cells, where the most concrete commonalities are stored. An example of Cluster is presented at the left of Figure~\ref{fig:Cluster}.

The Cells of a Cluster will process any new manifestation all in parallel, in a PDP fashion~\cite{rm14}, with the goal of updating the appropriate commonalities. To keep consistency between Cells, a consistency mechanism is necessary, so only the right Cells are updated. In the end, the goal would be that only one branch of the tree is updated, to keep the domains divided. However, these are implementation details that we will not address here.

Translating the SPPP principle to Clusters becomes a bit more nuanced, because in this case we have multiple Cells with multiple Projections. Thus, we need to decide which Projection will be the Cluster's one. In our case, we decided to take as Projection the most concrete representation of the input. The idea is that, as Cells process the input all in parallel, and then organises themselves to ensure consistency, there is going to be one Cell that has the Footprint with the highest similarity to the input, and thus that one is the right Projection of it. 

Now, up to this point we have been combining manifestations of the same type. However, there are multiple types of manifestations (i.e. image and sound), and many of them come at the same time. To be able to process those different types of manifestations independently, in order to build proper representations for each one, we can build multiple independent Clusters, one for each manifestation type. However, to be able to process the synchronicity of those different types of manifestations, we need to build a single Cluster that receives the synchronised inputs. A solution to this dichotomy is to build a Cluster that receives, as inputs, the concatenation of the representations made by those single type Clusters. This way, instead of building associations between individual manifestations (what we would achieve with the single Cluster), we will be building associations between already built representations. This allows the top Clusters to work with an extra level of abstraction. A group of Clusters connected between them in this fashion is what we call a \emph{Metacluster}. A representation of a Metacluster can be found in the right of Figure~\ref{fig:Cluster}.
To be able to build these Metaclusters, we first need to define another output of a Cluster: the \emph{Archetype}. An Archetype will be the most abstracted representation we have of a given input, and thus it will be a Footprint of the seed Cell. Then, a parent Cluster will receive the Archetypes of its children, and concatenate them to build its input.

When we translate the SPPP principle to a Metacluster we observe that this time it is quite straightforward: the Projection of a Metacluster will be the concatenation of the Projections of its leaf Clusters, as they conform an identification of the received input. However, it has a powerful effect: it allows to build sensorimotor loops. Let us take the example shown in Figure~\ref{fig:Cluster}. If we provide input only to the Image and Sound Clusters, and keep the Motor Cluster without input, the information of the input state will go up the hierarchy till the Top Cluster, where all the Footprints have information not only about the image and sound but also about the corresponding motor response. Thus, the self-projecting property of Footprints will force them to project also information about the motor response, that will be then projected also by their corresponding Cells, that will in turn be projected also by the Top Cluster. If we take this Projection and ask the Motor Cluster to process it, we will end with a Motor Cluster Projection that, after being decoded by the embodiment, will produce a motor response to the given input state.

Finally, the last level of our hierarchy will be the connection of multiple Metaclusters. This level is more interesting, since, at this level, we can start assigning functionalities to each Metacluster based on the Standard Model of the Mind~\cite{llr17}. So far, in our work we have identified three potential types of Metaclusters: the \emph{Motoperceptive Metacluster}, the \emph{Declarative Metacluster} and the \emph{Procedural Metacluster}.

The Motoperceptive Metacluster would be a Metacluster whose lower level Clusters receive as input the external signals coming from the embodiment. It covers all the signals, both input signals (perception) and output signals (motor), and thus it can control the whole body by itself. In its top Cluster, it associates the different inputs to the available outputs, or in agent theory terms, it associates the possible states to the possible actions, and thus it can contain a policy. In that sense, with this Metacluster alone we are able to build reactive agents, that is, those that can build sensorimotor abstractions. These agents will only be able to produce, for a given state, a fixed action, and thus they are quite limited, but with the right embodiment they can solve complex problems following a fixed policy. In that regard, this Metacluster will behave not only as Motor~\cite{llr17} and Perception~\cite{llr17}, but also in some sense as a Semantic Memory~\cite{llr17}. 

The Declarative Metacluster is a more complex Metacluster. Our current proposal is that this Metacluster would be conformed by a single Cluster, that receives inputs representing time. To do so, a Declarative Metacluster stores the last \emph{n} inputs received, and concatenates them to generate an episode, that later is fed to the Cluster of the Metacluster. Thus, this Cluster will build representations of episodes. In that regard, it will behave like a Declarative Memory~\cite{llr17}, or at least as an Episodic Memory~\cite{llr17}. This Metacluster was conceived to be connected to a Motoperceptive Metacluster, but it can be used also independently for certain tasks. When connected to a Motoperceptive Metacluster, it allows to build episodic agents, that is, those that can build episodic abstractions. These agents will be able to adapt to time context, and thus they will be able to produce, for a given state, different actions based on what happened before. However, they still follow a fixed policy and cannot reason.


Finally, the Procedural Metacluster is the most complex Metacluster, and it is still not fully defined. Thus, we do not have a firm proposal of what its inputs would be. The idea would be that this Metacluster will be placed between a Declarative Metacluster and a Motoperceptive Metacluster and will regulate their communication and potentially change the data with which they are working. In that sense, it will model though processes and will replicate them to solve problems. Thus, it should work like a Procedural Memory~\cite{llr17}, emerging the effects of a Working Memory in the interconnection between the Metaclusters. This Metacluster makes no sense without the other Metaclusters, although we do not discard that some problems could be solved with a Procedural Metacluster and the adequate embodiment.

\begin{figure}[tp]
    \centering
    \includegraphics[width=\columnwidth]{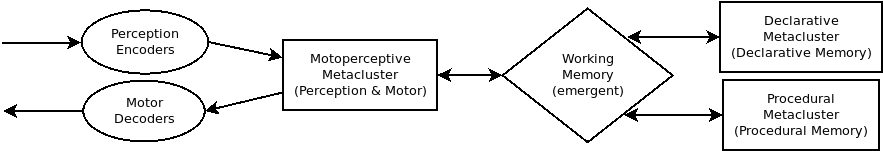}
    \caption{A generic example of Synthetic Cognition.}
    \label{fig:SC}
\end{figure}

The final goal is that an agent built with one of each kind of Metacluster will be a reasoning agent, that is, an agent able to build reasoning abstractions. The Motoperceptive Metacluster will deal with the low level, instantaneous actions and will work to identify and build representations of commonalities from the manifestations it receives from the embodiment. Then, the Declarative Metacluster will deal with time and will build representations of commonalities based on time, producing actions sensitive to context. Finally, the Procedural Metacluster will work with the other two to generate reasoning loops and complex behaviours, and will build abstractions of them. We call this construct of Metaclusters a \emph{Synthetic Cognition}, and an example is displayed at Figure~\ref{fig:SC}.

Finally, applying the SPPP principle to a Synthetic Cognition consist in setting as its Projection the Projection of the Motoperceptive Metacluster. This implies that, after all the processing, reasoning, etc..., the agent still produces signals that can be processed by the embodiment.

To end this section, we want to remark that the SPPP principle is fundamental to the Footprint definition for a basic reason: it transforms the representations into functions. This transformation turns the Footprint into a primitive that builds representations of commonalities and recognises them. Thus, we have a primitive that, with the consecutive hierarchical building, is able to build the functional elements of a cognitive architecture from the ground, in a \emph{minimal cognition} fashion~\cite{dkf06}.

\section{Conclusions}
In this paper we have focused on developing a modelling method to achieve Artificial General Intelligence, starting with our proposal of analysing the world from an information source perspective, and translating it to a computational framework. The framework presented has the potential to build classical cognitive architectures from the bottom-up, in a primitive-based fashion, constituting a huge novelty with respect the actual state-of-the-art. For future work, we would like to explore the potential of our framework to give an answer to the Fodor and Pylyshyn's Systematicity Challenge~\cite{fp88}.

\begin{credits}
\subsubsection{\ackname}
We want to thank Daniel Pinyol, Hector Antona and Pere Mayol for our insightful discussions about the topic.

\subsubsection{\discintname}
The authors have no competing interests to declare that are relevant to the content of this article.
\end{credits}

\bibliographystyle{splncs04}
\bibliography{biblio}

\end{document}